\documentclass[10pt,twocolumn,letterpaper]{article}

\usepackage{iccv}
\usepackage{times}
\usepackage{epsfig}
\usepackage{graphicx}
\usepackage{amsmath}
\usepackage{amssymb}

\usepackage{bbm}
\usepackage{multirow}
\usepackage{float}
\usepackage{placeins}

\usepackage{color}
\usepackage{colortbl}
\definecolor{blue}{rgb}{0,0,1}
\definecolor{ky}{rgb}{0.4,0.7,0}
\newcommand{\ky}[1]{\textcolor{black}{{#1}}}
\newcommand{\tuan}[1]{\textcolor{black}{{#1}}}

\definecolor{darkred}{rgb}{0.55, 0.0, 0.0}

\usepackage[breaklinks=true,bookmarks=false]{hyperref}

\iccvfinalcopy 

\ificcvfinal\fi

\begin{document}

\title{TemporalMaxer: Maximize Temporal Context with only Max Pooling \\ for Temporal Action Localization}

\author{Tuan N. Tang, Kwonyoung Kim, Kwanghoon Sohn* \\
	School of Electrical and Electronic Engineering \\
	Yonsel University \\
	{\tt\small \{tuantng, kyk12, khsohn\}@yonsei.ac.kr
}
}
\maketitle
\begin{abstract}

Temporal Action Localization (TAL) is a challenging task in video understanding that aims to identify and localize actions within a video sequence. 
\ky{Recent studies have} emphasized the importance of \ky{applying} long-term temporal context modeling (TCM) blocks \ky{to the extracted video clip features} such as \ky{employing complex self-attention mechanisms}. 
\ky{In this paper, we present the simplest method ever to address this task and argue that the extracted video clip features are already informative to achieve outstanding performance without sophisticated architectures.}
\ky{To this end, we} introduce TemporalMaxer, which minimizes long-term temporal context modeling while maximizing information from the extracted video clip features with a basic, parameter-free, and local region operating max-pooling block. 
\ky{Picking out} only the most critical information for adjacent and local clip embeddings, \ky{this block results} in a more efficient TAL model.
We demonstrate that TemporalMaxer outperforms other \ky{state-of-the-art methods} that utilize long-term TCM such as self-attention on various TAL datasets while requiring significantly fewer parameters and computational resources.
The code for our approach is publicly available at \url{https://github.com/TuanTNG/TemporalMaxer}.

\end{abstract}

\begin{figure*}[]
\begin{center}
\includegraphics[width=1.0\textwidth]{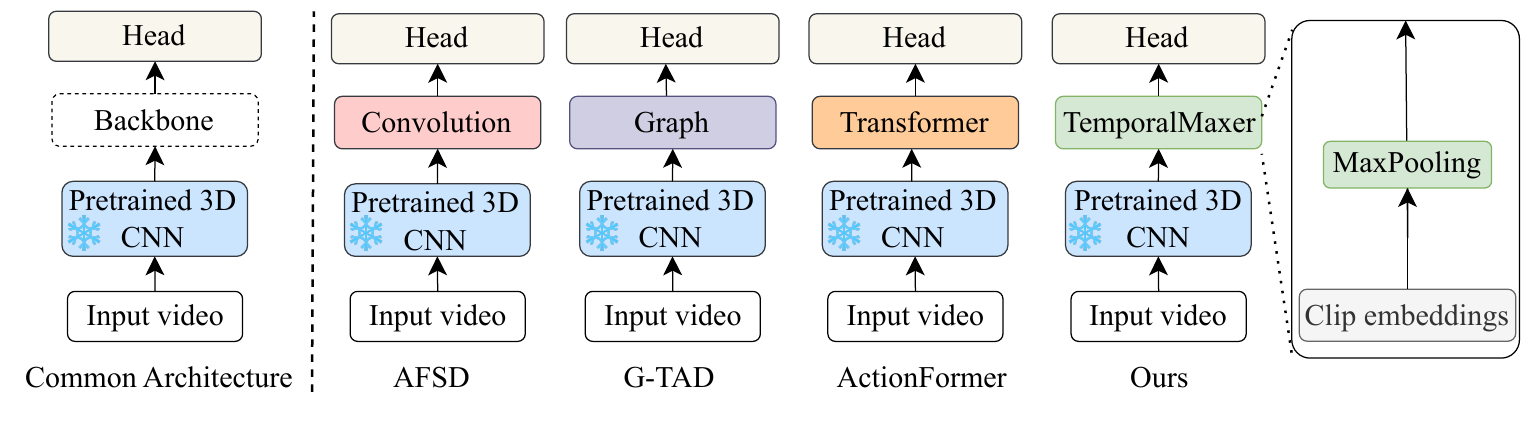}
\end{center}
\caption{Common architecture in Temporal Action Localization (TAL) and different temporal context modeling (TCM) blocks. Existing works have incorporated extensive parameters, high computational costs, and complex modules in a backbone such as 1D Convolutional layer in AFSD \cite{lin2021learning} to capture local temporal context, Graph \cite{kipf2016semi} in G-TAD \cite{xu2020g} and Transformers \cite{vaswani2017attention} in ActionFormer \cite{zhang2022actionformer} to model long-term temporal contexts. Our proposed method, termed as TemporalMaxer, for the first time exploits the potentials of strong features from pretrained 3D CNN by only utilizing a basic, parameter-free, local operating Max Pooling block. Our proposed method, the simplest backbone ever for TAL, maintains only the most critical information on adjacent and local clip embeddings. Combined with the large receptive field of deep networks, the whole model outperforms other works by a large margin on various datasets.}
\label{fig:tcm}
\end{figure*}
\let\thefootnote\relax\footnotetext{*Corresponding author}
\section{Introduction}
Temporal action localization (TAL), \ky{which aims to localize} action instances in time and \ky{assign} their categorical labels, {is a challenging but essential task} within the field of video comprehension. 
Various approaches have been proposed to address this task, such as action proposals \cite{lin2019bmn},  anchor windows \cite{long2019gaussian}, or dense prediction \cite{lin2021learning}. A widely accepted notion for improving the performance of TAL models is to integrate a component capable of capturing long-term temporal dependencies within the \ky{extracted video clip features}\cite{qing2021temporal, gao2020accurate, zhang2022actionformer, he2022glformer, liu2022end, nawhal2021activity, tan2021relaxed, wang2021temporal}. 
\tuan{Specifically,} \ky{the TAL model first employs pre-extracted features from a pre-trained 3D-CNN network, such as I3D \cite{carreira2017quo} and TSN \cite{wang2016temporal}, as an input. Then, an encoder, called backbone, encodes features to latent space, and the decoder, called head, predicts action instances as illustrated in Fig. \ref{fig:tcm}. To better capture long-term temporal dependencies, long-term temporal context modeling (TCM) blocks are incorporated into the backbone. Particularly, prior works \cite{xu2020g, zhao2021video} have employed Graph \cite{kipf2016semi}, or more complex modules such as Local-Global Temporal Encoder \cite{qing2021temporal} which uses a channel grouping strategy, or Relation-aware pyramid Network \cite{gao2020accurate} which exploits bi-directional long-range relations.}
\ky{Recently} there has been notable interest in the application of the self-attention \cite{vaswani2017attention} for long-term TCM \cite{zhang2022actionformer, liu2022end, zhao2021actionness, kang2022htnet}, resulting in surprising performance improvements.

\begin{figure}[hbt!]
\begin{center}
\includegraphics[width=1.0\linewidth]{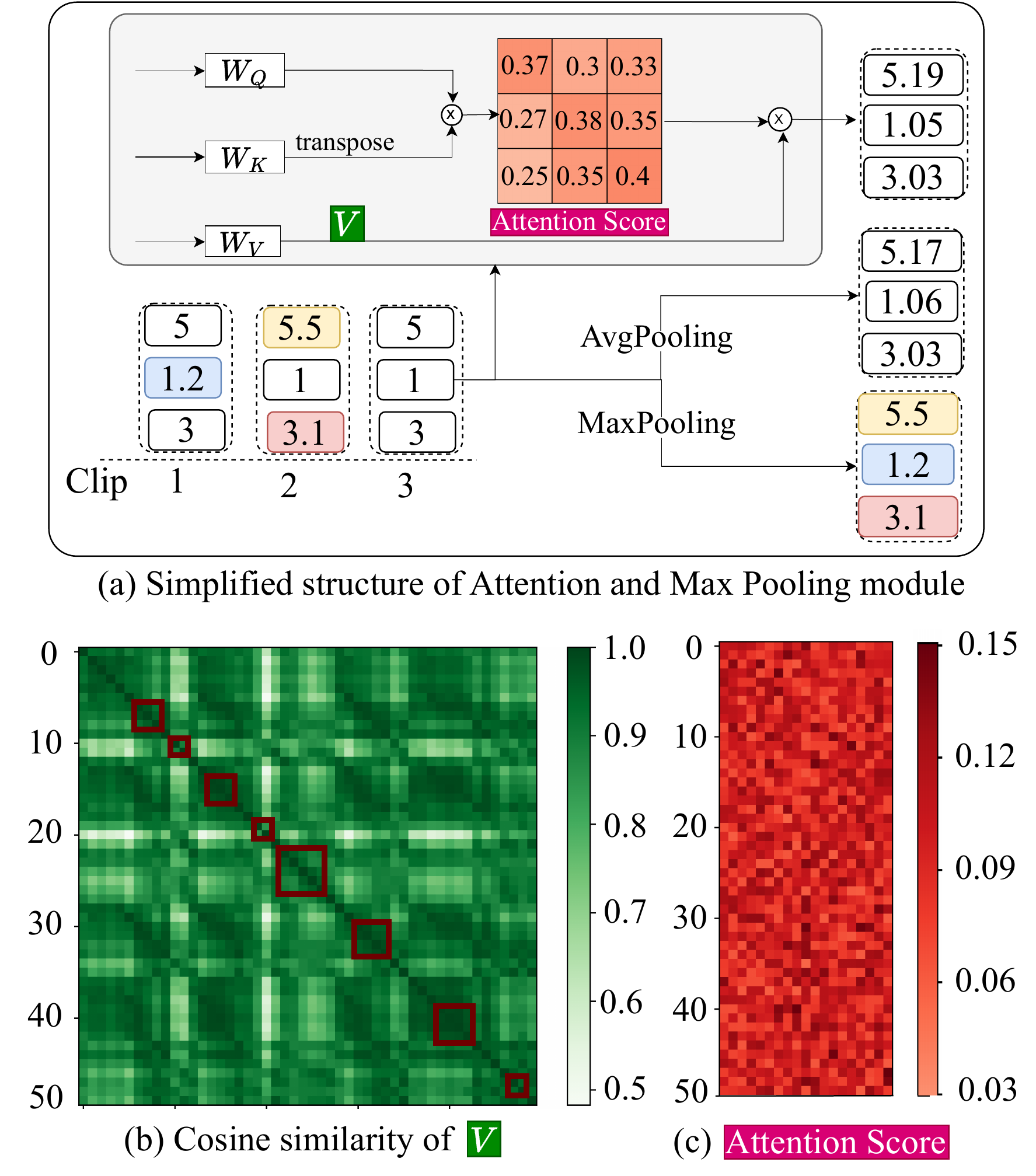}
\end{center}
\caption{\textbf{Comparison of effectiveness of Max Pooling over Transformer} \cite{vaswani2017attention}. 
(a) When the input clip embeddings are highly similar, given that the value features $V$ are highly similar and the attention score is distributed equally, the Transformer tends to average out the clip embeddings as done in Average Pooling. 
In contrast, Max Pooling can retain the most crucial information about adjacent clip embeddings and further remove redundancy in the video.
(b) Cosine similarity matrix(bottom-left) of value features in self-attention in ActionFormer \cite{zhang2022actionformer} where the \textcolor{darkred}{red boxes} exhibit the action intervals.
(c) Attention score(bottom-right) after training the ActionFormer\cite{zhang2022actionformer}. }
\label{fig:similarity}
\end{figure}



\ky{Long-term TCM} \tuan{in the backbone} \ky{can help the model to capture long-term temporal dependencies between frames, which can be useful in identifying complex actions that may unfold over longer periods of time or heavily overlapping actions.}
Despite their performance improvement on benchmarks, the inference speed and the effectiveness of the existing approaches are rarely considered. While \cite{zhang2022actionformer} introduced and pursued the minimalist design, their backbone is still limited to the transformer architecture requiring \tuan{expensive parameters} and computations.

Recently, \cite{yu2022metaformer} proposed general architecture abstracted from transformer. 
Motivated by the success of recent approaches replacing attention module with MLP-like modules\cite{tolstikhin2021mlp} or Fourier Transform\cite{lee2021fnet}, they deemed the essential of those modules as \textit{token mixer} which aggregates information among tokens. In turn, they came to propose \textit{PoolFormer}, equiped with extremely minimized token mixer which replaces the exhausting attention module with very simple pooling layer. 

In this paper, motivated by their proposal, we focus on extreme minimization of the backbone network by concentrating on and maximizing the information from the extracted video clip features in a short-term perspective rather than employing exhausting long-term encoding process. 

The transformer architecture leading state-of-the-art performance in the machine translation task \cite{vaswani2017attention} and computer vision areas \cite{dosovitskiy2020image} have inspired recent works in TAL \cite{zhang2022actionformer, liu2022end, zhao2021actionness, kang2022htnet}. From the perspective of long-term TCM, the property that calculates attention weights for the \tuan{long} input sequence has led to recent progress in the TAL. However, such long-term consideration comes at a price of high computation costs \tuan{and the effectiveness of those approaches have not yet been carefully analyzed}. 


We argue the essential properties of the video clip features that have not been fully exploited to date. 
Firstly, the video clips exhibit a high redundancy which leads to a high similarity of the pre-extracted features as demonstrated in Fig. \ref{fig:similarity}. It raises the question of the effectiveness of employing self-attention or graph methods for long-range TCM. Unlike in other domains such as \tuan{machine translation task} where input tokens exhibit distinctiveness, the input clip embeddings in TAL frequently exhibit a high degree of similarity. Consequently, as depicted in Fig. \ref{fig:similarity}, the self-attention recently employed in TAL tend to average the embeddings of clips within the attention scope, losing temporally local minute changes by redundant similar frames under the long-term temporal context modeling.
We argue that only certain information within clip embeddings is relevant to the action context of TAL, whereas the remainder of the information is similar across adjacent clips. Therefore, an optimal TCM must be capable of preserving the most discriminative features of clip embeddings that carry the essential information.

To this end, we aim to \ky{propose simple yet effective} TCM in a straightforward manner. 
We argue that the TCM can retain the simplest architecture while maximizing informative features extracted from 3D CNN.
Max Pooling \cite{boureau2010theoretical} is deemed the most fitting block for this purpose. 

Our proposed method, TemporalMaxer, presents the simplest architecture ever for this task gaining much faster inference speed.  
Our finding suggests that the pre-trained feature extractor already possess great potential and with those feature, short-term TCM solely can benefit the performance for this task.

Extensive experiments prove the superiority and effectiveness of the proposed method, showing state-of-the-art performance in terms of both accuracy and speed for TAL on various challenging datasets including THUMOS \cite{idrees2017thumos}, EPIC-Kitchens 100 \cite{damen2020rescaling}, MultiTHUMOS \cite{yeung2018every}, and MUSES \cite{liu2021multi}.


\section{Related Work}
\textbf{Temporal Action Localization} (TAL). Two-Stage and Single-Stage methods are used in TAL to detect actions in videos. Two-Stage methods first generate possible action proposals and classify them into actions. The proposals are generated through anchor windows \cite{escorcia2016daps, buch2017sst, heilbron2016fast}, detecting action boundaries \cite{lin2018bsn, gong2020scale, zhao2020bottom}, graph representation \cite{bai2020boundary, xu2020g}, or Transformers \cite{wang2021temporal, chang2021augmented, tan2021relaxed}.
Single-stage TAL performs both action proposal generation and classification in a single pass, without using a separate proposal generation step. The pioneering work \cite{qing2021temporal} developed anchor-based single-stage TAL using convolutional networks, inspired by a single-stage object detector \cite{redmon2016you, liu2016ssd}. \ky{Meanwhile,} \cite{lin2021learning} proposed an anchor-free single-stage model with a saliency-based refinement module.

\textbf{Long-term Temporal Context Modeling (TCM)}. Recent studies have emphasized the necessity of long-term TCM to improve model performance. Long-term TCM helps the model capture long-term temporal dependencies between frames, which can be useful in identifying intricate actions that span over extended timeframes or heavily overlapping actions. Prior work has addressed long-term TCM using Relation-aware pyramid Network \cite{gao2020accurate}, Multi-Stage CNN \cite{farha2019ms}, Temporal Context Aggregation Network \cite{qing2021temporal}. Recent works \cite{xu2020g, zhao2021video} employ Graph \cite{kipf2016semi} where each video clip feature represents a node in a graph as long-term TCM. More recently, Transformer \cite{vaswani2017attention} demonstrates an outstanding capacity to capture long-range dependency of the input sequence in the machine translation tasks. Thus, it is a natural fit for Temporal Action Localization (TAL) where each video clip embedding represents a token. Therefore, recent studies \cite{zhang2022actionformer, liu2022end, zhao2021actionness, kang2022htnet} have employed Transformers as a long-term TCM.

Our approach, TemporalMaxer, belongs to the single-stage TAL model that utilizes a state-of-the-art ActionFormer \cite{zhang2022actionformer} as the baseline for comparison. Similar to ActionFormer, TemporalMaxer follows a minimalistic design of sequence labeling where every moment is classified, and their corresponding action boundaries are regressed. 
The main difference is that we avoid exhausting attention between clips from long-term timeframes which can unintentionally flatten minute information among the crowd of similar frames, but keep the minute information in short-term manner.


\section{Method}
\subsection{Problem Statement}
\begin{figure*}[t]
\centering
\begin{center}
\resizebox{.8\textwidth}{!}{%
\includegraphics[width=1.0\linewidth]{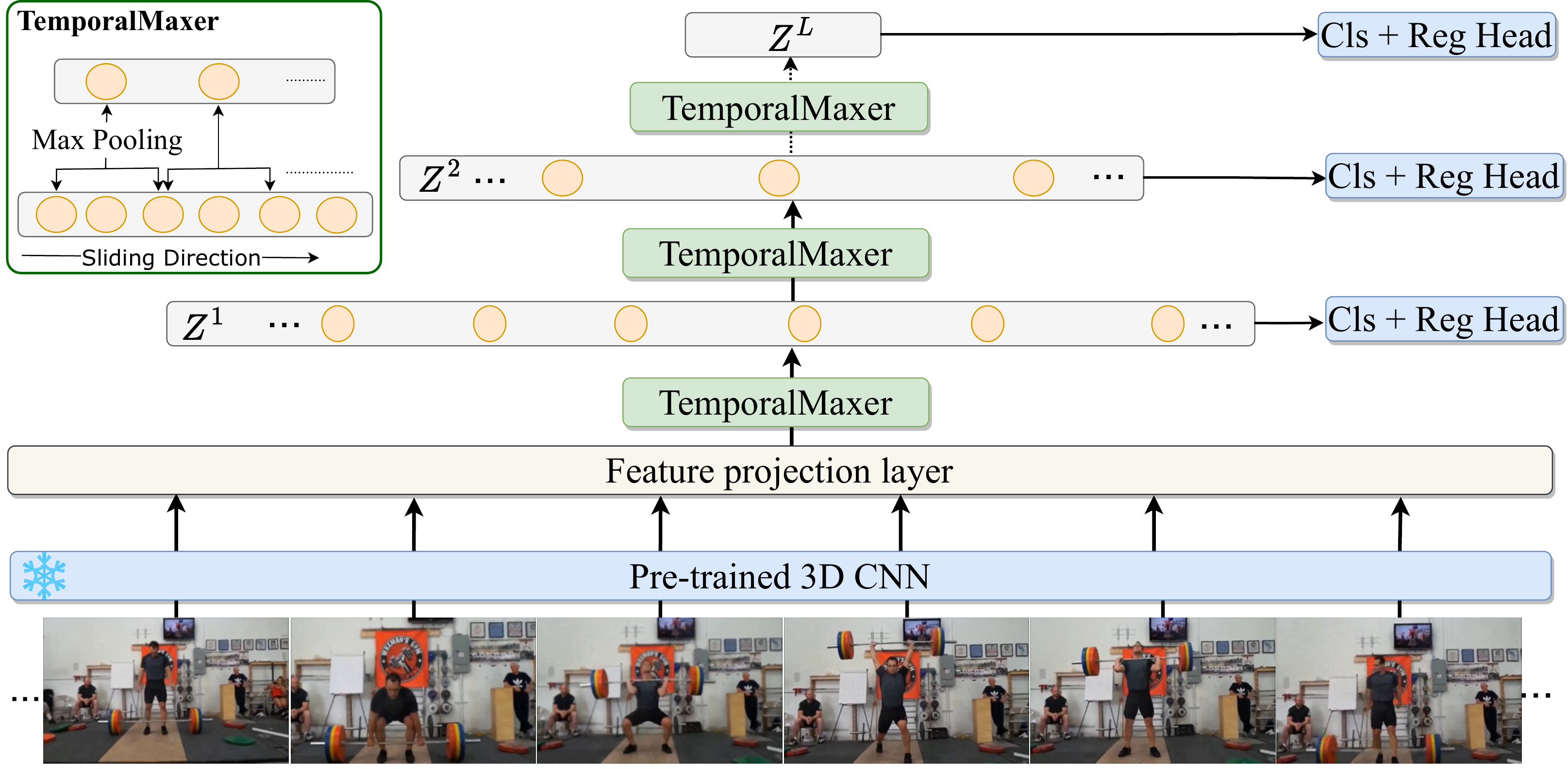}}
\end{center}
\caption{Overview of TemporalMaxer. The proposed method utilizes Max Pooling as a Temporal Context Modeling block applied between temporal feature pyramid levels to maximize informative features of high similarity clip embedding. Specifically, it first extracts features of every clip using pre-trained 3D CNN. After that, the backbone encodes clip features to form a multi-scale feature pyramid. The backbone consists of 1D convolutional layers and TemporalMaxer layers. Finally, a lightweight classification and regression head decodes the feature pyramid to action candidates for every input moment.}
\label{fig:architecture}
\end{figure*}

\textbf{Temporal Action Localization}. Assume that an untrimmed video $X$ can be represented by a set of feature vectors $X = \lbrace{x_1, x_2, . . . , x_T }\rbrace$, where the number of discrete time steps $t = \lbrace{1, 2, . . . , T}\rbrace$ may vary depending on the length of the video. The feature vector $x_t$ is extracted from a pre-trained 3D convolutional network and represents a video clip at a specific moment $t$. The aim of TAL is to predict a set of action instances $\Psi = \{\psi_1, \psi_2, \ldots, \psi_N \}$ based on the input video sequence $X$, where $N$ is the number of action instances in $X$. Each action instance $\psi_n$ consists of ($s_n$, $e_n$, $a_n$) where $s_n$, $e_n$, and $a_n$ are starting time, ending time, and associated action label $a_n$ respectively, $s_n \in [1, T]$, $e_n \in [1, T]$, $s_n < e_n$, and the action label $a_n$ belongs to the pre-defined set of $C$ categories.


\subsection{TemporalMaxer}

\textbf{Action Representation}. We follow the anchor-free single-stage representation \cite{lin2021learning, zhang2022actionformer} for an action instance. Each moment is classified into the background or one of $C$ categories, and regressed the onset and offset based on the current time step of that moment. Consequently, the prediction in TAL is formulated as a sequence labeling problem.
\begin{equation}
X=\left\{x_1, x_2, \ldots, x_T\right\} \rightarrow \hat{\Psi}=\left\{\hat{\psi}_1, \hat{\psi}_2, \ldots, \hat{\psi}_T\right\}
\end{equation}
At the time step $t$, the output $\hat{\psi}_t = (o_t^s, o_t^e, c_t)$ is defined as following:
\begin{itemize}
\item $o_t^s > 0$ and $o_t^e > 0$ represent the temporal intervals between the current time step $t$ and the onset and offset of a given moment, respectively.
\item Given $C$ action categories, the action probability $c_t$ can be considered as a set of $c_t^i$ which is a probability for action $i^{th}$, where $1\leq i \leq C$.
\end{itemize}
The predicted action instance at time step $t$ can be retrieved from $\hat{\psi}_t = (o_t^s, o_t^e, c_t)$ by:
\begin{equation}
a_t=\arg \max \left(c_t\right), \quad s_t=t-o_t^s, \quad e_t=t+o_t^e
\end{equation}

\textbf{Architecture overview}. The overall architecture is depicted in Fig. \ref{fig:architecture}. Our proposed method aims to learn to label every input moment by $f\left(X\right) \rightarrow \hat{\Psi}$ with $f$ is a deep learning model. $f$ follows an encoder-decoder design and can be decomposed as $e \circ d$. The encoder here is the backbone, and the decoder is the classification and regression head. $e: X \rightarrow Z$ learn to encode the input video feature $X$ into latent vector $Z$, and $d: Z \rightarrow \hat{\Psi}$ learns to predict labels for every input moment. To effectively capture actions transpiring at various temporal scales, we also adopt a multi-scale feature pyramid representation, which is denoted as $Z = \{Z^1, Z^2, . . . , Z^L\}$.


\textbf{Encoder design}
The input feature $X$ is first encoded into multi-scale temporal feature pyramid $Z = \{Z^1, Z^2, . . . , Z^L\}$ using encoder $e$. The encoder $e$ simply contains two 1D convolutional neural network layers as feature projection layers, followed by $L-1$ Temporal Context Modeling (TCM) blocks to produce feature pyramid $Z$. Formally, the feature projection layers are described as:
\begin{equation} \label{eq:projection}
    X_{p} = E_2(E_1(\text{Concat}(X)))
\end{equation}

The input video feature sequence $X = \lbrace{x_1, x_2, . . . , x_T }\rbrace$, where $x_i \in \mathbb{R}^{1 \times D_{in}}$,  is first concatenated in the first dimension and then fed into two feature projection modules $E_1$, and $E_2$ in equation \ref{eq:projection}, resulting in the projected feature $X_{p} \in \mathbb{R}^{T \times D}$ with D-dimensional feature space. Each projection module comprises one 1D-convolutional neural network layer, followed by Layer Normalization \cite{ba2016layer}, and ReLU \cite{agarap2018deep}. We simply assign $Z^1 = X_p$ as the first feature in $Z$. Finally, the multi-scale temporal feature pyramid $Z$ is encoded by TemporalMaxer:
\begin{equation} \label{eq:zl}
    Z^l = \text{TemporalMaxer}(Z^{l-1}).
\end{equation}
Here: TemporalMaxer is Max Pooling and employed with stride 2, $Z^l \in \mathbb{R}^{\frac{T}{2^{l-1}} \times D}$, $2 <= l <= L$. It is worth noting that ActionFormer \cite{zhang2022actionformer} employs Transformer \cite{vaswani2017attention} as a TCM block where each clip feature at the moment $t$ represents a token, our proposed method adopts only Max Pooling \cite{boureau2010theoretical} as TCM block.

\textbf{Decoder Design}
The decoder $d$ learns to predict sequence labeling, $\hat{\Psi}=\left\{\hat{\psi}_1, \hat{\psi}_2, \ldots, \hat{\psi}_T\right\}$, for every moment using multi-scale feature pyramid $Z = \{Z^1, Z^2, . . . , Z^L\}$. The decoder adopts a lightweight convolutional neural network and consists of classification and regression heads. Formally, the two heads are defined as:

\begin{equation} \label{eq:cls}
    C_l = \mathcal{F}_c(E_4(E_3(Z^l)))
\end{equation}
\begin{equation} \label{eq:reg}
    O_l = \text{ReLU}(\mathcal{F}_o(E_6(E_5(Z^l))))
\end{equation}
Here, $Z^l \in \mathbb{R}^{\frac{T}{2^{l-1}} \times C}$ is the latent feature of level $l$, $C_l=\{c_0, c_{2^{l-1}},...,c_T\} \in \mathbb{R}^{\frac{T}{2^{l-1}} \times C}$ denotes the classification probability with $c_i \in \mathbb{R}^{C}$, and $O_l = \{(o_0^s, o_0^e), (o_{2^{l-1}}^s, o_{2^{l-1}}^e),...,(o_T^s, o_T^e)\} \in \mathbb{R}^{\frac{T}{2^{l-1}} \times 2}$ is the onset and offset prediction of input moment $\{0, 2^{l-1},..., T\}$. $E$ denotes the 1D convolution followed by Layer Normalization and ReLU activation function. $\mathcal{F}_c$ and $\mathcal{F}_o$ are both 1D convolution. Note that all the weights of the decoder are shared between the different features in the multi-scale feature pyramid $Z$.

\textbf{Learning Objective}. The model predicts $\hat{\psi}_t = (o_t^s, o_t^e, c_t)$ for every moment of the input $X$. Following the baseline \cite{zhang2022actionformer}, the Focal Loss \cite{lin2017focal} and DIoU loss \cite{zheng2020distance} are employed to supervise classification and regression outputs respectively. The overall loss function is defined as:
\begin{equation}
\mathcal{L}_{total}=\sum_t\left(\mathcal{L}_{cls}+\mathbbm{1}_{c_t} \mathcal{L}_{reg}\right) / T_{+}
\end{equation}
where $\mathcal{L}_{reg}$ denotes regression loss and is applied only when the indicator function, $\mathbbm{1}_{c_t}$, indicates that the current time step $t$ is a positive sample. $T_{+}$ is the number of positive samples. $\mathcal{L}_{cls}$ is $C$ way classification loss. The loss function $\mathcal{L}_{total}$ is applied to all levels on the output of multi-scale feature pyramid $Z$ and averaged across all video samples during training.


\section{Experimental Results}
In this section, we show that our proposed method, TemporalMaxer, demonstrates the outstanding results achieved across a variety of challenging datasets, namely THUMOS \cite{idrees2017thumos}, EPIC-Kitchens 100 \cite{damen2020rescaling}, MultiTHUMOS \cite{yeung2018every}, and MUSES \cite{liu2021multi}. These datasets are recognized as standard benchmarks in the Temporal Action Localization task. Our approach surpasses the state-of-the-art baseline, ActionFormer \cite{zhang2022actionformer}, in each dataset, showcasing its superior performance compared to other works.

\textbf{Evaluation Metric}. We employ a widely-used evaluation metric for TAL known as the mean average precision (mAP) calculated at various temporal intersections over union (tIoU). tIoU is the intersection over union between two temporal windows, i.e., the 1D Jaccard index. We report the mAP scores for all action categories based on the given tIoU thresholds, and further report an averaged mAP value across all tIoU thresholds.

\textbf{Training Details}. To ensure a fair and unbiased comparison, we employed the experimental setup of the robust baseline model, ActionFormer \cite{zhang2022actionformer}. This setup included various components such as decoder design $d$, non-maximum suppression (NMS) hyper-parameters in the post-processing stage, data augmentation, learning rate, feature extraction, and the number of feature pyramid level $L$. The sole variation in our study was the substitution of the Transformer block in ActionFormer with the Max Pooling block. All experiments are conducted with a kernel size of 3 for all TCM blocks. The subsequent ablation  will thorough analysis of the effects of varying kernel sizes. During training, the input feature length is kept constant at 2304, corresponding to approximately 5 minutes of video on both THUMOS14 and MultiTHUMOS datasets, roughly 20 minutes on the EPIC-Kitchens 100 dataset, and approximately 45 minutes on MUSES. Additionally, Model EMA \cite{huang2017snapshot} and gradient clipping techniques are employed, consistent with those used in \cite{zhang2022actionformer}, to promote training stability.

\subsection{Results on THUMOS14}

\textbf{Dataset}. THUMOS14 dataset \cite{idrees2017thumos} contains 200 validation videos and 213 testing videos with 20 action classes. Following previous work \cite{lin2019bmn, lin2018bsn, xu2020g, zhao2020bottom, zhang2022actionformer}, we trained the model using validation videos and measured the performance on testing videos.

\textbf{Feature Extraction}. Following \cite{zhang2022actionformer, zhao2020bottom}, we extract the features of THUMOS14 dataset using two-stream I3D \cite{carreira2017quo} pre-trained on Kinetics \cite{kay2017kinetics}. 16 consecutive frames are fed into I3D pre-trained network with a sliding window of stride 4. The extracted feature is collected after the last fully connected layer and has 1024-D feature space. After that, the two-stream features are further concatenated (2048-D) and utilized as the input of the model.

\textbf{Results}. We compare the performance
evaluated on the THUMOS14 dataset \cite{idrees2017thumos} with state-of-the-art methods. TemporalMaxer demonstrates remarkable performance, achieving an average mAP of 67.7\% mAP, outperforming all previous approaches, both single-stage, and two-stage methods, by a significant margin, with a 1.1\% increase in mAP at tIoU=0.4.
Especially, TemporalMaxer surpasses all recent methods that utilize long-term TCM blocks including self-attention such as TadTR \cite{liu2022end}, HTNet \cite{kang2022htnet}, TAGS \cite{nag2022proposal}, GLFormer \cite{he2022glformer}, ActionFormer \cite{zhang2022actionformer}, or Graph-based like G-TAD \cite{xu2020g}, VSGN \cite{zhao2021video}, or complex module including Local-Global Temporal Encoder \cite{qing2021temporal}.

Moreover, the comparisons between our method and other approaches show that the proposed method not only exhibits outstanding performance but also is efficient in terms of inference speed. Specifically, our method only takes 50 ms on average to fully process an entire video on THUMOS. It is 1.6x faster than the ActionFormer baseline and 3.9x faster than TadTR. However, in section \ref{section:ablation}, we show that our model only takes 10.4 ms for forward time. It means that the rest 39.6 ms is NMS time, causing the most time-consuming. In the later section \ref{section:ablation}, we show that the forward time and the backbone time of our model are 2.9x and 8.0x faster than ActionFormer, respectively.

\begin{table*}[]
\centering
\resizebox{.9\textwidth}{!}{%
\begin{tabular}{cccccccccc}
\hline
\multirow{2}{*}{Type}         & \multirow{2}{*}{Model}                                                                   & \multirow{2}{*}{Feature}                         & \multicolumn{6}{c}{tIoU$\uparrow$}         & \multirow{2}{*}{time(ms) $\downarrow$} \\ \cline{4-9} 
                              &                                                                                          &                                                  & 0.3           & 0.4           & 0.5           & 0.6           & 0.7           & Avg.                 \\ \hline
\multirow{14}{*}{Two-Stage}   & BMN \cite{lin2019bmn}                                                   & TSN \cite{wang2016temporal}     & 56.0          & 47.4          & 38.8          & 29.7          & 20.5          & 38.5 & 483* \\
                              & DBG \cite{lin2020fast}                                                  & TSN \cite{wang2016temporal}     & 57.8          & 49.4          & 39.8          & 30.2          & 21.7          & 39.8 & --- \\
                              & G-TAD \cite{xu2020g}                                                    & TSN \cite{wang2016temporal}     & 54.5          & 47.6          & 40.3          & 30.8          & 23.4          & 39.3 & 4440* \\
                              & BC-GNN \cite{bai2020boundary}                                           & TSN \cite{wang2016temporal}     & 57.1          & 49.1          & 40.4          & 31.2          & 23.1          & 40.2 & --- \\
                              & TAL-MR \cite{zhao2020bottom}                                            & I3D \cite{carreira2017quo}      & 53.9          & 50.7          & 45.4          & 38.0          & 28.5          & 43.3 & \textgreater644* \\
                              & P-GCN \cite{zeng2019graph}                                              & I3D \cite{carreira2017quo}      & 63.6          & 57.8          & 49.1          & ---           & ---           & ---  & 7298* \\
                              & P-GCN \cite{zeng2019graph} +TSP \cite{alwassel2021tsp} & R(2+1)1 D \cite{tran2018closer}                  & 69.1          & 63.3          & 53.5          & 40.4          & 26.0          & 50.5 & --- \\
                              & TSA-Net \cite{gong2020scale}                                            & P3D \cite{qiu2017learning}      & 61.2          & 55.9          & 46.9          & 36.1          & 25.2          & 45.1 & --- \\
                              & MUSES \cite{liu2021multi}                                               & I3D \cite{carreira2017quo}      & 68.9          & 64.0          & 56.9          & 46.3          & 31.0          & 53.4 & 2101* \\
                              & TCANet \cite{qing2021temporal}                                          & TSN \cite{wang2016temporal}     & 60.6          & 53.2          & 44.6          & 36.8          & 26.7          & 44.3 & --- \\
                              & BMN-CSA \cite{sridhar2021class}                                         & TSN \cite{wang2016temporal}     & 64.4          & 58.0          & 49.2          & 38.2          & 27.8          & 47.7 & --- \\
                              & ContextLoc \cite{zhu2021enriching}                                      & I3D \cite{carreira2017quo}      & 68.3          & 63.8          & 54.3          & 41.8          & 26.2          & 50.9 & --- \\
                              & VSGN \cite{zhao2021video}                                               & TSN \cite{wang2016temporal}     & 66.7          & 60.4          & 52.4          & 41.0          & 30.4          & 50.2 & --- \\
                              & RTD-Net \cite{tan2021relaxed}                                           & I3D \cite{carreira2017quo}      & 68.3          & 62.3          & 51.9          & 38.8          & 23.7          & 49.0 & \textgreater211* \\ 
                              & Disentangle \cite{zhu2022learning}                                           & I3D \cite{carreira2017quo} & 72.1          & 65.9          & 57.0          & 44.2          & 28.5          & 53.5 & --- \\
                              & SAC \cite{yang2022structured}                                           & I3D \cite{carreira2017quo}      & 69.3          & 64.8          & 57.6          & 47.0          & 31.5          & 54.0 & --- \\
                              \cline{2-10}
\multirow{7}{*}{Single-Stage} & A²Net \cite{yang2020revisiting}                                         & I3D \cite{carreira2017quo}      & 58.6          & 54.1          & 45.5          & 32.5          & 17.2          & 41.6 & 1554* \\
                              & GTAN \cite{long2019gaussian}                                            & P3D \cite{qiu2017learning}      & 57.8          & 47.2          & 38.8          & ---           & ---           & ---  & --- \\
                              & PBRNet \cite{liu2020progressive}                                        & I3D \cite{carreira2017quo}      & 58.5          & 54.6          & 51.3          & 41.8          & 29.5          & ---  & --- \\
                              & AFSD \cite{lin2021learning}                                             & I3D \cite{carreira2017quo}      & 67.3          & 62.4          & 55.5          & 43.7          & 31.1          & 52.0 & 3245* \\
                              & TAGS \cite{nag2022proposal}                                                 & I3D \cite{carreira2017quo}  & 68.6          & 63.8          & 57.0          &  46.3         & 31.8          &  52.8 & ---\\
                              & HTNet \cite{kang2022htnet}                                                 & I3D \cite{carreira2017quo}   & 71.2          & 67.2          & 61.5          &  51.0         & 39.3          &  58.0 & ---\\
                              & TadTR \cite{liu2022end}                                                 & I3D \cite{carreira2017quo}      & 74.8          & 69.1          & 60.1          & 46.6          & 32.8          & 56.7  & 195* \\
                              & GLFormer \cite{he2022glformer}                                                 & I3D \cite{carreira2017quo}      & 75.9 & 72.6 & 67.2 & 57.2 & 41.8 & 62.9 & ---\\
                              & AMNet \cite{liu2022end}                                                 & I3D \cite{carreira2017quo}      & 76.7          & 73.1          & 66.8          & 57.2          & 42.7          &  63.3 & ---\\
                              & ActionFormer \cite{zhang2022actionformer}                               & I3D \cite{carreira2017quo}      & 82.1          & 77.8          & 71.0          & 59.4          & 43.9          & 66.8 & 80 \\
                              & ActionFormer \cite{zhang2022actionformer} + GAP \cite{nag2022post} & I3D \cite{carreira2017quo}      & 82.3          & ---          & 71.4          & ---       &  44.2          & 66.9  & \textgreater 80 \\
                              \cline{2-10}
                              & Our (TemporalMaxer) & I3D \cite{carreira2017quo}      & \textbf{82.8} & \textbf{78.9} & \textbf{71.8} & \textbf{60.5} & \textbf{44.7} & \textbf{67.7} &  \textbf{50} \\ \cline{2-10}   
\end{tabular}
}
\caption{The results obtained on the THUMOS14 dataset \cite{idrees2017thumos} are presented for various tIoU thresholds, with the average mAP calculated in the range [0.3:0.7:0.1]. The top-performing results are highlighted in bold. The time(ms) is the average inference time for one video, without extracting features from 3D CNN and including the post-processing step, such as NMS. We measure the inference time using a single GeForce GTX 1080 Ti GPU. Results indicated with * are taken from \cite{liu2022end} which are measured using Tesla P100 GPU, a much more powerful GPU than the 1080 Ti. In comparison to early works, including both one-stage and two-stage methods, and those utilizing long-term TCM, TemporalMaxer achieves superior performance in both mAP and inference speed.}
\label{table:sota_thumos}
\end{table*}

\subsection{Results on EPIC-Kitchens 100}
\textbf{Dataset}. The EPIC-Kitchens 100 dataset \cite{damen2020rescaling} is a comprehensive collection of egocentric action videos, featuring 100 hours of footage from 700 sessions that document cooking activities in a variety of kitchens. Additionally, EPIC-Kitchens 100 is three times larger in terms of total video hours and more than ten times larger in terms of action instances (averaging 128 per video) when compared to THUMOS14. These videos are recorded from a first-person perspective, resulting in significant camera motion, and represent a novel challenge for TAL research.

\textbf{Feature Extraction}. Following previous work \cite{zhang2022actionformer, lin2019bmn, xu2020g}, we extract the videos feature using SlowFast network \cite{feichtenhofer2019slowfast} pre-trained on EPICKitchens \cite{damen2020rescaling}. We utilized a 32-frame input sequence with a stride of 16 to generate a set of 2304-D features. These features were then fed as input to our model.

\textbf{Result}. Tab. \ref{table:sota_epic} shows our results. TemporalMaxer demonstrates notable performance on the EPIC-Kitchens 100 dataset, achieving an average mAP of 24.5\% and 22.8\% for verb and noun, respectively. The superiority of our approach is further confirmed by a large margin over a strong and robust baseline, ActionFormer \cite{zhang2022actionformer}, with an average improvement of 1.0\% mAP for verb and 0.9\% mAP for noun. Again, TemporalMaxer outperforms other methods that utilize long-term TCM including self-attention \cite{zhang2022actionformer} or Graph \cite{xu2020g}. These results provide empirical evidence of the effectiveness of the simplest backbone in advancing the state-of-the-art on this challenging task.

\begin{table*}[]
\centering
\resizebox{.6\textwidth}{!}{%
\begin{tabular}{cccccccc}
\hline
\multirow{2}{*}{Task} & \multirow{2}{*}{Method}                                    & \multicolumn{6}{c}{tIoU}                                                                      \\ \cline{3-8}
                      &                                                            & 0.1           & 0.2           & 0.3           & 0.4           & 0.5           & Avg           \\ \hline
\multirow{4}{*}{Verb} & BMN \cite{lin2019bmn, damen2020rescaling} & 10.8          & 9.8           & 8.4           & 7.1           & 5.6           & 8.4           \\
                      & G-TAD \cite{xu2020g}                      & 12.1          & 11.0          & 9.4           & 8.1           & 6.5           & 9.4           \\
                      & ActionFormer \cite{zhang2022actionformer} & 26.6          & 25.4          & 24.2          & 22.3          & 19.1          & 23.5          \\
                      & Our (TemporalMaxer)                                        & \textbf{27.8} & \textbf{26.6} & \textbf{25.3} & \textbf{23.1} & \textbf{19.9} & \textbf{24.5} \\ \cline{2-8}
\multirow{4}{*}{Noun} & BMN \cite{lin2019bmn, damen2020rescaling} & 10.3          & 8.3           & 6.2           & 4.5           & 3.4           & 6.5           \\
                      & G-TAD \cite{xu2020g}                      & 11.0          & 10.0          & 8.6           & 7.0           & 5.4           & 8.4           \\
                      & ActionFormer \cite{zhang2022actionformer} & 25.2          & 24.1          & 22.7          & 20.5          & 17.0          & 21.9          \\
                      & Our (TemporalMaxer)                                        & \textbf{26.3} & \textbf{25.2} & \textbf{23.5} & \textbf{21.3} & \textbf{17.6} & \textbf{22.8} \\ \hline
\end{tabular}
}
\caption{The performance of our proposed method on the EPIC-Kitchens 100 dataset \cite{damen2020rescaling} is evaluated using various tIoU thresholds. The average mAP is reported over a range of tIoU thresholds [0.1:0.5:0.1]. The top-performing methods are highlighted in bold. Our proposed method outperforms the other methods significantly.}
\label{table:sota_epic}
\end{table*}

\subsection{Results on MUSES}
\textbf{Dataset}. The MUSES dataset \cite{liu2021multi} is a collection of 3,697 videos, with 2,587 for training and 1,110 for testing. MUSES has 31,477 action instances over a duration of 716 video hours with 25 action classes, designed to facilitate multi-shot analyses making the dataset challenging.

\textbf{Feature Extraction}. We directly employ the pre-extracted feature provided by \cite{liu2021multi}. The feature is extracted using a pre-trained I3D network\cite{carreira2017quo} on the Kinetics dataset \cite{kay2017kinetics} using only RGB stream, resulting in 1024-D feature space for a video clip embedding.

\textbf{Result}
Tab. \ref{table:sota_muses} shows the results on MUSES dataset. Our method significantly outperforms other works that employ long-term TCM such as G-TAD \cite{xu2020g}, Ag-Trans \cite{zhao2021actionness}, and ActionFormer \cite{zhang2022actionformer} at every tIoU threshold. Notably, TemporalMaxer improves 1.0 mAP at tIoU=0.7. On average, we achieve 27.2 mAP, which is 1.0 mAP higher than the previous approaches, demonstrating the robustness of our method. It is worth noting that we implemented the ActionFormer \cite{zhang2022actionformer} on the MUSES dataset using the code provided by the authors.

\begin{table}[]
\centering
\begin{tabular}{ccccc}
\hline
\multirow{2}{*}{Method}                   & \multicolumn{4}{c}{tIoU}                \\ \cline{2-5}
                                          & 0.3  & 0.5   & 0.7  & Avg  \\ \hline
BU-TAL \cite{zhao2020bottom}              & 12.9 & 9.2   & 5.9  & 9.4  \\
G-TAD \cite{xu2020g}                      & 19.1 & 11.1  & 4.7  & 11.4 \\
P-GCN \cite{zeng2019graph}                & 19.9 & 13.1  & 5.4  & 13.0 \\
MUSES \cite{liu2021multi}                 & 25.9 & 18.9 & 10.6 & 18.6 \\
Ag-Trans \cite{zhao2021actionness}        & 24.8 & 19.4 & 10.9 & 18.6 \\
ActionFormer \cite{zhang2022actionformer} & 35.9 & 26.9 & 15.2 & 26.2 \\ \hline
Our (TemporalMaxer)                       & \textbf{36.7} & \textbf{27.8} & \textbf{16.2} & \textbf{27.2} \\ \hline
\end{tabular}
\caption{We report mAP at different tIoU thresholds [0.3, 0.5, 0.7] and the average mAP in [0.3:0.1:0.7] on MUSES dataset \cite{liu2021multi}. All methods used the same I3D features. Our method outperforms concurrent works by a large margin.}
\label{table:sota_muses}
\end{table}

\subsection{Results on MultiTHUMOS}
\textbf{Dataset}. The MultiTHUMOS dataset \cite{yeung2018every} is a densely labeled extension of THUMOS14, consisting of 413 sports videos with 65 distinct action classes. The dataset presents a significant increase in the average number of distinctive action categories per video, compared to THUMOS14. As such, it poses a greater challenge for TAL than THUMOS. While MultiTHUMOS are being used in action detection benchmark \cite{dai2021pdan, dai2022ms}, a novel approach for action detection, PointTAD \cite{tan2022pointtad}, utilizes the TAL evaluation metric to assess the completeness of predicted action instances. Given that TAL and action detection share the same setting in terms of input features and annotations, we evaluate the performance of our model on MultiTHUMOS and compare it against the state-of-the-art action detection methods \cite{dai2021pdan, tirupattur2021modeling, dai2022ms, tan2022pointtad}, and strong baseline ActionFormer \cite{zhang2022actionformer}.


\textbf{Feature Extraction}. We only utilize RGB stream as input for I3D network \cite{carreira2017quo} pre-trained on Kinetics \cite{kay2017kinetics}, following \cite{tan2022pointtad}, to extract features for MultiTHUMOS. The I3D pre-trained network is fed with 16 sequential frames through a sliding window with a stride of 4. The feature is extracted from the final fully connected layer, resulting in a 1024-D feature space that serves as input for the model.

\textbf{Result}. Tab. \ref{table:sota_multihumos} provides a comparison of our performance on the MultiTHUMOS dataset \cite{yeung2018every} with recent state-of-the-art methods. Specifically, our method surpasses the prior work \cite{tan2022pointtad} that utilizes TransFormer as feature encoding by a large margin, 6.4\% mAP on average. Moreover, TemporalMaxer improves the robust baseline, ActionFormer \cite{zhang2022actionformer}, by 2.4\% mAP at tIoU=0.7, and 1.3\% mAP on average. It should be noted that we utilized the code provided by the authors to implement ActionFormer \cite{zhang2022actionformer} on the MultiTHUMOS dataset, and the results \cite{dai2021pdan, tirupattur2021modeling, dai2022ms} are taken from \cite{tan2022pointtad}.

\begin{table}[]
\centering
\resizebox{.43\textwidth}{!}{%
\begin{tabular}{ccccc}
\hline
\multirow{2}{*}{Method}                                               & \multicolumn{4}{c}{tIoU}                                      \\ \cline{2-5}
                                                                      & 0.2           & 0.5           & 0.7           & Avg           \\ \hline
PDAN \cite{dai2021pdan}                              & ---           & ---           & ---           & 17.3          \\
MLAD \cite{tirupattur2021modeling}                   & ---           & ---           & ---           & 14.2          \\
MS-TCT \cite{dai2022ms}                              & ---           & ---           & ---           & 16.2          \\
PointTAD \cite{tan2022pointtad}                      & 39.7          & 24.9          & 12.0          & 23.5          \\
ActionFormer \cite{zhang2022actionformer} & 46.4          & 32.4          & 15.0          & 28.6          \\ \hline
Our (TemporalMaxer)                                                   & \textbf{47.5} & \textbf{33.4} & \textbf{17.4} & \textbf{29.9} \\ \hline
\end{tabular}}%
\caption{Comparison with the state-of-the-art methods on the MultiTHUMOS dataset. We report the results at different tIoU thresholds [0.2, 0.5, 0.7] and average mAP in [0.1:0.9:0.1].}
\label{table:sota_multihumos}
\end{table}

\subsection{Ablation Study}
\label{section:ablation}
We perform various ablation studies to verify the effectiveness of TemporalMaxer. To better understand what is the effective component for TCM, we gradually replace the  Max Pooling with other blocks such as convolution, subsampling, and Average Pooling. Furthermore, we evaluate numerous kernel sizes of Max Pooling. Note that all experiments in this section are conducted on the train and validation set of the THUMOS14 dataset.

\begin{table*}[]
\centering
\resizebox{.9\textwidth}{!}{%
\begin{tabular}{cccccccc}
\hline
\multirow{2}{*}{TCM}& \multicolumn{3}{c}{tIoU}  & \multirow{2}{*}{GMACs $\downarrow$} & \multirow{2}{*}{\#params (M) $\downarrow$} & \multirow{2}{*}{time (ms) $\downarrow$} & \multirow{2}{*}{backbone time (ms) $\downarrow$}\\ \cline{2-4}
& 0.5 & 0.7 & Avg.\\ \hline
Conv \cite{lin2021learning} (Our Impl) & 62.8 & 37.1 & 59.4 & 45.6 & 30.5 & 16.3 & 9.0\\
Subsampling  & 64.3 & 37.7 & 61.0 & \textbf{16.2} & 7.1 & 10.4 & 2.5\\
Average Pooling & 66.1 & 39.4 & 63.2 & 16.4 & 7.1 & 10.4 & 2.5\\
Transformer \cite{vaswani2017attention} & 71.0 & 43.9 & 66.8 & 45.3 & 29.3 & 30.5 & 20.1 \\ \hline
TemporalMaxer & \textbf{71.8} & \textbf{44.7} & \textbf{67.7} & 16.4 & \textbf{7.1} & \textbf{10.4} &  \textbf{2.5} \\ \hline
\end{tabular}
}
\caption{Ablation studies about different TCM blocks on THUMOS14. Inference times are measured using an input video with 2304 clip embeddings, a 5 minutes video, on a GeForce GTX 1080 Ti GPU without post-processing (NMS) and pre-extracted features step.}
\label{table:ablation_block}
\end{table*}

\textbf{Effective of TemporalMaxer}. Tab. \ref{table:ablation_block} presents the results of other blocks other than Max Pooling.
Motivated by PoolFormer \cite{yu2022metaformer} that replaces the computationally intensive and highly parameterized attention module with the most basic block in deep learning, the pooling layer. Our studies started by first questioning the most straightforward approach to leverage the potential of the extracted features from 3D-CNN for the TAL task. PoolFormer retains the structure of Transformer such as FFN \cite{rosenblatt1961principles, rumelhart1985learning}, residual connection \cite{he2016deep}, and Layer Normalization \cite{ba2016layer} because the tokens have to be encoded by the Poolformer itself. However, in TAL the features from 3D CNN have already been pre-extracted and contain useful information. Therefore, we posit that there will be a straightforward block that are more efficient than Transformer/PoolFormer, does not require much computational as well as parameters, and be able to effectively exploit the pre-extracted features.

Our ablation study starts to employ a 1-D convolution module as an alternative to Transformer \cite{vaswani2017attention} for TCM block to see how many mAP drop. The result is reported in the first row of Tab. \ref{table:ablation_block}. As expected, the convolution layer decreases by 7.4\% mAP compared with the ActionFormer baseline. This deduction can be explained for two reasons. First, given the video clip features that are informative but highly similar, and the convolution weight is fixed after training, consequently the convolution operation cannot retain the most informative features of local clip embeddings. Secondly, the convolution layer introduces the most parameters which tend to overparameterize the model. We argue that given the informative features extracted from the pretrained 3D CNN, the model should not contain too many parameters, which may lead to overfitting and thus less generalization. To further concrete our thought, we replace the Transformer with a parameter-free operation, subsampling technique, in which features at even indexes are kept and odd indexes are removed to simulate stride 2 of TCM. Surprisingly, this none-parameter operation achieves higher results than the convolution layer, shown in the second row of Tab. \ref{table:ablation_block}. This finding proves that TCM block may not need to contain many parameters, and the features from pretrained 3D CNN are informative and is the potential for TAL.

However, subsampling is prone to losing the most crucial information as only half of the video clip embeddings are kept after a TCM block. Thus, we replace Transformer with Average Pooling with kernel size 3 and stride 2. As expected, Average Pooling improves the result by 2.2 \% mAP on average compared with the subsampling. It is because this operation does not drop any clip embedding which helps to retain the crucial information. However, the Average Pooling averages the nearby features, thus the crucial information of adjacent clips is reduced. That is why Average Pooling decreases by 3.6\% mAP compared with Transformer.

The ablation study with Average Pooling suggests that the most important information of clip embeddings should be maintained. For that reason, we employ Max Pooling as TCM. The result is provided in the last row of Tab. \ref{table:ablation_block}. Max Pooling achieves the highest results at every tIoU threshold and significantly outperforms the strong and robust baseline, ActionFormer. To clarify, TemporalMaxer effectively highlights only critical information from nearby clips and discards the less important information. This result suggests that the feature from pretrained 3D CNN are informative and can be effectively utilized for the TAL model without complex modules like prior works.

Our method, TemporalMaxer, results in the simplest model ever for TAL task that contains minimalist parameters and computational cost for the TAL model. TemporalMaxer is effective at modeling temporal contexts, which outperforms the robust baseline, ActionFormer, with 2.8x fewer GMACs and 3x faster inference speed. Especially, when comparing only the backbone time, our proposed method only takes 2.5 ms which is incredibly 8.0x faster than ActionFormer backbone \cite{zhao2021actionness}, 20.1 ms. 

\textbf{Different Values of kernel size.}
We make an ablation with the kernel size of TemporalMaxer 3, 4, 5, 6 and the average mAPs are 67.7, 67.1, 66.8, 65.7, respectively. Our model achieved the highest performance with a kernel size of 3, while the lowest performance was observed with a kernel size of 6. This decrease in performance can be attributed to the corresponding loss of information during training.
The results obtained for kernel sizes 3, 4, and 5 are not very sensitive to the kernel size. This suggests that these kernel sizes can effectively capture the relevant temporal information for the task at hand.



\section{Conclusion}
In this paper, we propose an extremely simplified approach, TemporalMaxer for temporal action localization task. To minimize the structure, we explore the way to simply maximize the underlying information in the video clip features from pre-trained 3D-CNN. To this end, with a basic, non-parametric and temporally local operating max-pooling block which can effectively and efficiently keep the local minute changes among the sequential similar input images.
We achieve competitive performance to other state-of-the-art methods with sophisticated, parametric and long-term temporal context modeling models. 




{\small
\bibliographystyle{ieee_fullname}
\bibliography{egbib}
}

\end{document}